%% file: 2018-SIBGRAPI-IrisSeg_main.tex
\documentclass[10pt,conference]{IEEEtran}

\usepackage{times}
\usepackage{epsfig}
\usepackage{graphicx}
\usepackage{amsmath}
\usepackage{amssymb}

\usepackage[acronym, shortcuts]{glossaries}
\usepackage{url}
\usepackage{footmisc}
\usepackage{multirow}
\usepackage{balance}
\usepackage[dvipsnames]{xcolor}
\usepackage[utf8]{inputenc}
\usepackage{booktabs}
\usepackage{dblfloatfix}
\usepackage{cite}\usepackage[colorlinks=true,pagebackref=false,urlcolor=black,linkcolor=black,citecolor=black]{hyperref} 

\newacronymstyle{long-short-br}
{%
  \GlsUseAcrEntryDispStyle{long-short}%
}%
{%
  \GlsUseAcrStyleDefs{long-short}%
}
\setacronymstyle{long-short-br}

\newcommand{\figI}{0.121}

\newacronym{dcnn}{DCNN}{Deep Convolutional Neural Network}
\newacronym{cnn}{CNN}{Convolutional Neural Network}
\newacronym{biosec}{BioSec}{BioSec}
\newacronym{casiai3}{CasiaI3}{CASIA-Iris-Interval-v3}
\newacronym{casiat4}{CasiaT4}{CASIA-Iris-Thousand-v4}
\newacronym{iitd1}{IITD-1}{IITD Iris Image Database 1.0}
\newacronym{ntdame}{NTDame}{Notredame 0405 Iris}
\newacronym{nice1}{NICE.I}{Noisy Iris Challenge Evaluation - Part~I}
\newacronym{nice2}{NICE.II}{Noisy Iris Challenge Evaluation - Part~II}
\newacronym{creyeiris}{CrEye-Iris}{Cross-Spectral Iris/Periocular}
\newacronym{miche1}{MICHE-I}{Mobile Iris Challenge Evaluation I}
\newacronym{fcn}{FCN}{Fully Convolutional Network}
\newacronym{gan}{GAN}{Generative Adversarial Network}
\newacronym{cgan}{CGAN}{Conditional Generative Adversarial Network}
\newacronym{osiris}{OSIRISv4.1}{Open Source Iris Recognition System Version~4.1}
\newacronym{irisseg}{IRISSEG}{Iris Segmentation Framework}
\newacronym{nir}{NIR}{near-infrared}
\newacronym{vis}{VIS}{visible}
\newacronym{mrs}{MRS}{Maximum Radial Suppression}
\newacronym{mtm}{MTM}{Markovian Texture Model}
\newacronym{ocac}{OCAC}{Optical Correlation based Active Contours}
\newacronym{hcnn}{HCNN}{Hierarchical Convolutional Neural Network}
\newacronym{mfcn}{MFCN}{Multi-scale Fully Convolutional Network}
\newacronym{f1}{F1}{F-Measure}
\newacronym{macc}{Mean Acc.}{Mean Accuracy}
\newacronym{iou}{IoU}{Intersection over Union}
\newacronym{tp}{TP}{True Positive}
\newacronym{fp}{FP}{False Positive}
\newacronym{tn}{TN}{True Negative}
\newacronym{fn}{FN}{False Negative}
\newacronym{prd}{PRD}{Periocular Region Detection}
\newacronym{roi}{ROI}{Region of Interest}
\newacronym{encdec}{ED}{Encoder-Decoder}

\usepackage{cite}
\usepackage{amsthm}

\usepackage{algorithmic}

\usepackage{array}

\ifCLASSOPTIONcompsoc
  \usepackage[caption=false,font=normalsize,labelfont=sf,textfont=sf]{subfig}
\else
  \usepackage[caption=false,font=footnotesize]{subfig}
\fi
\usepackage{url}

\begin{document}
%
\title{Robust Iris Segmentation Based on Fully Convolutional~Networks and Generative \\Adversarial Networks}

\newif\iffinal
\finaltrue
\newcommand{\jemsid}{58}


\iffinal

\author{Cides S. Bezerra\IEEEauthorrefmark{1}, Rayson Laroca\IEEEauthorrefmark{1}, Diego R. Lucio\IEEEauthorrefmark{1}, Evair Severo\IEEEauthorrefmark{1},\\Lucas F. Oliveira\IEEEauthorrefmark{1}, Alceu S. Britto Jr.\IEEEauthorrefmark{2} and David Menotti\IEEEauthorrefmark{1}\\
\IEEEauthorrefmark{1}Federal University of Paran\'a (UFPR), Curitiba, PR, Brazil\\
\IEEEauthorrefmark{2}Pontifical Catholic University of Paran\'a (PUCPR), Curitiba, PR, Brazil\\
{\tt\small \{csbezerra,rblsantos,drlucio,ebsevero,lferrari,menotti\}@inf.ufpr.br alceu@ppgia.pucpr.br}}


%

\else
  \author{SIBGRAPI paper ID: \jemsid \\ }
\fi

\maketitle
\thispagestyle{empty}

\begin{abstract}
The iris can be considered as one of the most important biometric traits due to its high degree of uniqueness.
Iris-based biometrics applications depend mainly on the iris segmentation whose suitability is not robust for different  environments such as \gls*{nir} and \gls*{vis} ones.
In this paper, two approaches for robust iris segmentation based on \glspl*{fcn} and \glspl*{gan} are described.
Similar to a common convolutional network, but without the fully connected layers (i.e., the classification layers), an \gls*{fcn} employs at its end a combination of pooling layers from different convolutional layers.
Based on the game theory, a \gls*{gan} is designed as two networks competing with each other to generate the best segmentation.
The proposed segmentation networks achieved promising results in all evaluated datasets (i.e., \acs*{biosec}, \acs*{casiai3}, \acs*{casiat4}, \acs*{iitd1}) of \gls*{nir} images and (\acs*{nice1}, \acs*{creyeiris} and \acs*{miche1}) of \gls*{vis} images in both non-cooperative and cooperative domains, outperforming the baselines techniques which are the best ones found so far in the literature, i.e., a new state of the art for these datasets.
Furthermore, we manually labeled 2,431 images from \acs*{casiat4}, \acs*{creyeiris} and \acs*{miche1} datasets, making the masks available for research purposes.
\end{abstract}

\IEEEpeerreviewmaketitle

\input{introduction.tex}
\input{related.tex}
\input{proposed.tex}
\input{experiments.tex}
\input{results.tex}
\balance
\input{conclusion.tex}

\section*{Acknowledgments}
This work was supported by grants from the National Council for Scientific and Technological Development~(CNPq) (\#~428333/2016-8, \#~313423/2017-2 and \#~307277/2014-3) and the Coordination for the Improvement of Higher Education Personnel~(CAPES). The Titan Xp GPU used for this research was donated by the NVIDIA Corporation. 


{\bibliographystyle{IEEEtran}
\bibliography{egbib}
}

\end{document}

%% file: introduction.tex
\section{Introduction}
\label{sec:introduction}
\glsresetall

The identification of individuals based on their biological and behavioral characteristics has a higher degree of reliability compared to other means of identification, such as passwords or access cards. 
Several characteristics of the human body can be used for person recognition (e.g., face, signature, fingerprints, iris, sclera, retina, voice, etc.)~\cite{Jain50years_2016}.
The characteristics present in the iris make it one of the most representative and safe biometric modalities.
This circular diaphragm forming the textured portion of the eye is capable of distinguishing individuals with a high degree of uniqueness~\cite{wildes1997iris,jain_Ross_Prabhakar_2004}.

As described in~\cite{Jillela20154}, an automated biometric system for iris recognition is composed of four main steps: (i)~image acquisition, (ii)~iris segmentation, (iii)~normalization and (iv)~feature extraction and matching.
The segmentation consists of locating and isolating the iris from other regions (e.g., the sclera,  surrounding skin regions, etc.), therefore it is the most critical and challenging step of the system.
Incorrect segmentation usually affects the subsequent steps, impairing the system performance~\cite{Rattani_2017_Survey_ocular_biometrics}.

Over the last decade, many approaches have been employed for iris segmentation, such as those based on edge detection~\cite{Liu_Bowyer_Kevin_Flynn_2005_WAIAT}, Hough transform~\cite{Proenca_Alexandre_2006_VISP}, active contours~\cite{Ouabida_2017_iris,Shah_Ross_2009_TIFS}, integro-differential equation~\cite{Tan_He_Sun_BestNICE-I_2010}, \gls*{mrs}~\cite{Podder_Khan_Khan_Rahman_Ahmed_2015_ICCCI}, \glspl*{mtm}~\cite{Haindl_Krupicka_2015}, and \glspl*{cnn}~\cite{Jalilian_2017_domain_adaptation, Liu_Zhang_2016_cnn_iris} (see~Section~\ref{sec:related_work} for more details).

Leveraging the advent of \glspl*{cnn} we propose two approaches for iris segmentation task. The first is based on a \gls*{fcn}~\cite{teichmann2016multinet} and the second one is based on a \gls*{gan}~\cite{isola2017imagetoimage}.
\glspl*{fcn} are used for segmentation in many different tasks since medical image analysis to aerospace image analysis~\cite{DBLP:journals/corr/abs-1802-01445, Roth2018}, while \gls*{gan} is a young approach to semantic segmentation, which has outperformed the state of the art~\cite{DBLP:journals/corr/LucCCV16}. 

The proposed \gls*{fcn} and \gls*{gan} iris segmentation approaches outperform three existing frameworks in the largest benchmark datasets found in the literature.
There are two main contributions in this paper:
(i) two \gls*{cnn}-based approaches that work well for \gls*{nir} and \gls*{vis} images in both cooperative (highly controlled) and non-cooperative environments; 
and~(ii)~$2{,}431$ new manually labeled masks from images of three existing iris datasets\footnote{The new masks are publicly available to the research community at \url{http://web.inf.ufpr.br/vri/databases/iris-segmentation-annotations/}.} (see Section~\ref{subsec:datasets}).

The remainder of this paper is organized as follows: 
we briefly review related work in Section~\ref{sec:related_work}. 
In Section~\ref{sec:proposed_approach}, the proposed approaches used for iris segmentation are described. 
Section~\ref{sec:experiments} presents the datasets, evaluation protocol and baselines used in the experiments. 
We report and discuss the results in Section~\ref{sec:results}. 
Conclusions are given in Section~\ref{sec:conclusion}.	

%% file: related.tex
\section{Related Work}
\label{sec:related_work}

In this section, we briefly review relevant studies in the context of iris segmentation, which use from conventional image processing to deep learning techniques.
For other studies on iris segmentation, please refer to~\cite{jan2017segmentation,DeMarsico_2018}.

Jillela and Ross~\cite{Jillela_Ross_2016_chapter7} presented an overview of classical approaches, evaluation methods and challenges related to iris segmentation in both \gls*{nir} and \gls*{vis} images. Daugman's study~\cite{daugman:1993} is considered the pioneer in iris segmentation.
The integro-differential operator was used to approximate the boundary of the inner and outer iris, generating the central coordinates and both pupil and iris radius.

Liu et al.~\cite{Liu_Bowyer_Kevin_Flynn_2005_WAIAT} first detected the inner boundary of the iris and then the outer boundary.
In addition, noisy pixels were eliminated based on their high/low-intensity level.
Proen\c{c}a and Alexandre~\cite{Proenca_Alexandre_2006_VISP} used the Fuzzy K-means algorithm to classify each pixel as belonging to a group, considering its coordinates and intensity distribution.
Then, they applied the Canny edge detector in the image with the grouped pixels, creating an edge map.
Finally, the inner and outer iris boundaries are detected by the circular Hough transform.

Shah and Ross~\cite{Shah_Ross_2009_TIFS} performed iris segmentation through Geodesic Active Contours, combining energy minimization with active contours based on curve evolution.
The pupil is detected from a binarization and both inner and outer iris boundaries are approximated using the Fourier series coefficients.
 
The winning approach of the \gls*{nice1}, proposed by Tan et al.~\cite{Tan_He_Sun_BestNICE-I_2010}, removes the reflection points using adaptive thresholding and bilinear interpolation.
Region growing based on clustering and integro-differential constellation segments the iris.
Podder et al.~\cite{Podder_Khan_Khan_Rahman_Ahmed_2015_ICCCI} applied an \gls*{mrs} technique to noise removal. 
Moreover, they applied the Canny edge detector and Hough transform to detect iris boundaries.

Haindl \& Krupi\v{c}ka~\cite{Haindl_Krupicka_2015} detected the iris using the Daugman's operator~\cite{daugman:1993} and removed the eyelids employing a third-order polynomial mean and standard deviation estimates. 
Adaptive thresholding and \gls*{mtm} were used to remove iris reflection.
Ouabida et al.~\cite{Ouabida_2017_iris} applied the \gls*{ocac}, that uses the Vander Lugt correlator algorithm, to detect the iris and pupil contours through spatial filtering.

Liu et al.~\cite{Liu_Zhang_2016_cnn_iris} proposed two approaches called \glspl*{hcnn} and \glspl*{mfcn} to perform a dense prediction of the pixels using sliding windows, merging shallow and deep layers.

At present, \glspl*{cnn} are being employed to solve many computer vision problems with impressive results being obtained in several areas such as biometrics, medical imaging and security systems~\cite{Ahuja_2017_CNN_ocularBiometric, dumoulin_2016_guideConvolution, severo2018benchmark}.
Teichmann et al.~\cite{teichmann_2016_multinet} proposed a \gls*{cnn} architecture, called MultiNet, to joint detection, classification and semantic segmentation. Inspired by the great results reported in their work, we apply the segmentation decoder of the MultiNet to the iris segmentation context, as detailed in Section~\ref{sub_sec:iris_segmentation}.

%% file: proposed.tex
\section{Proposed Approach}
\label{sec:proposed_approach}

This section describes the proposed approach and it is divided into two subsections, one for iris location and one for iris segmentation.

\subsection{Iris Detection}
\label{sub_sec:eye_detection}

The datasets used in this work have many different sizes, and just resizing the images would generate a distortion in the iris format. In order to avoid this distortion, we first performed the \gls*{prd}.

YOLO~\cite{redmon2016yolo} is a real-time object detection system, which regards detection as a regression problem.
As great advances were recently attained through models inspired by YOLO~\cite{laroca2018robust,severo2018benchmark}, we decided to fine-tune it for \gls*{prd}. However, as we want to detect only one class (i.e., the iris), we chose to use a smaller model, called Fast-YOLO\footnote{For training Fast-YOLO we used the weights pre-trained on ImageNet, available at \url{https://pjreddie.com/darknet/yolo/}.}~\cite{redmon2016yolo}, which uses fewer convolutional layers than YOLO and fewer filters in those layers. The Fast-YOLO’s architecture is shown in Table~\ref{tab:fast_yolo2}.

\begin{table}[!htb]
	\centering	
	\caption{Fast-YOLO network used for iris detection.}
	\label{tab:fast_yolo2}	
	\vspace{-1mm}
	\resizebox{0.98\columnwidth}{!}{	
		\begin{tabular}{@{}cccccc@{}}
			\toprule
			\multicolumn{2}{c}{\textbf{Layer}} & \textbf{Filters} & \textbf{Size} & \textbf{Input} & \textbf{Output} \\ \midrule
			$0$ & conv & $16$ & $3 \times 3 / 1$ & $416 \times 416 \times 1 / 3$ & $416 \times 416 \times 16$ \\
			$1$ & max &  & $2 \times 2 / 2$ & $416 \times 416 \times 16$ & $208 \times 208 \times 16$ \\
			$2$ & conv & $32$ & $3 \times 3 / 1$ & $208 \times 208 \times 16$ & $208 \times 208 \times 32$ \\
			$3$ & max &  & $2 \times 2 / 2$ & $208 \times 208 \times 32$ & $104 \times 104 \times 32$ \\
			$4$ & conv & $64$ & $3 \times 3 / 1$ & $104 \times 104 \times 32$ & $104 \times 104 \times 64$ \\
			$5$ & max &  & $2 \times 2 / 2$ & $104 \times 104 \times 64$ & $52 \times 52 \times 64$ \\
			$6$ & conv & $128$ & $3 \times 3 / 1$ & $52\times 52 \times 64$ & $52 \times 52 \times 128$ \\
			$7$ & max &  & $2 \times 2 / 2$ & $52 \times 52 \times 128$ & $26 \times 26 \times 128$ \\
			$8$ & conv & $256$ & $3 \times 3 / 1$ & $26 \times 26 \times 128$ & $26 \times 26 \times 256$ \\
			$9$ & max &  & $2 \times 2 / 2$ & $26 \times 26 \times 256$ & $13 \times 13 \times 256$ \\
			$10$ & conv & $512$ & $3 \times 3 / 1$ & $13 \times 13 \times 256$ & $13 \times 13 \times 512$ \\
			$11$ & max &  & $2 \times 2 / 1$ & $13 \times 13 \times 512$ & $13 \times 13 \times 512$ \\
			$12$ & conv & $1024$ & $3 \times 3 / 1$ & $13 \times 13 \times 512$ & $13 \times 13 \times 1024$ \\
			$13$ & conv & $1024$ & $3 \times 3 / 1$ & $13 \times 13 \times 1024$ & $13 \times 13 \times 1024$ \\
			$14$ & conv & $30$ & $1 \times 1 / 1$ & $13 \times 13 \times 1024$ & $13 \times 13 \times 30$ \\
			$15$ & detection &  &  &  &  \\ \bottomrule
		\end{tabular}} \,
\end{table}

The \acs*{prd} network was trained using the images, without any preprocessing, and the coordinates of the \gls*{roi} as inputs. The annotations provided by Severo et al.~\cite{severo2018benchmark} were used as ground truth. We applied a small padding in the detected patch to increase the chance that the iris is entirely within the \gls*{roi}. Afterward, we enlarged the \acs*{roi} to a square form with width and height that are power of $2$.

By default, only objects detected
with a confidence of $0.25$ or higher are returned by Fast-YOLO~\cite{redmon2016yolo}. We consider only the
detection with the largest confidence in cases where more
than one iris region is detected, since there is always
only one region annotated in the evaluated datasets. If no region is detected, the next stage (iris segmentation) is
performed on the image in its original size.

In our previous work on sclera segmentation~\cite{lucio2018fully}, this same approach was used for iris detection.

\subsection{Iris Segmentation}
\label{sub_sec:iris_segmentation}

We chose \gls*{fcn} and \gls*{gan} for iris segmentation since they presented good results in other segmentation applications~\cite{lucio2018fully}.
These results can be explained by the fact that \gls*{fcn} has no fully connected layer which generally causes loss of spatial information, while the representations embodied by the pair of networks in a \gls*{gan} model (the generator and the discriminator) are able to capture the statistical distribution of training data, making possible less reliance on huge, well-balanced, and well-labelled datasets.

\begin{figure}[!htb]
	\begin{center}
		\includegraphics[scale=0.25]{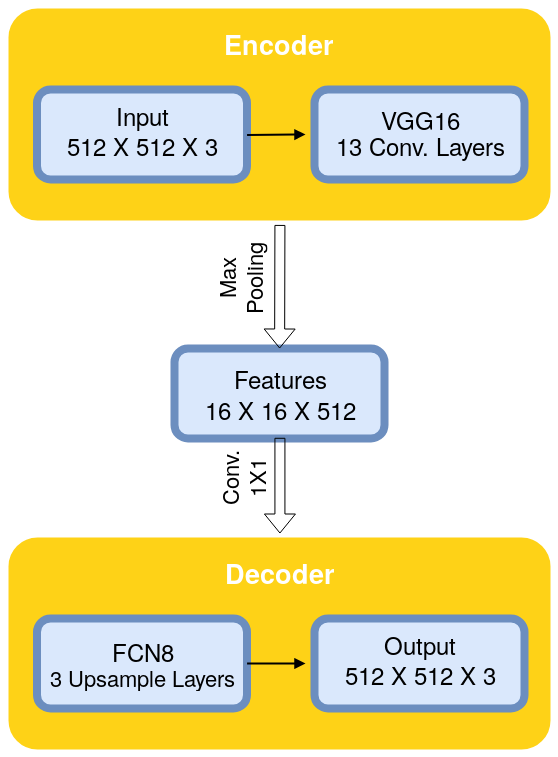}
	\end{center}
	\caption{\gls*{fcn} architecture for iris segmentation.}
	\label{fig:architecture_fcn}
\end{figure}

\subsubsection{Fully Convolutional Networks (FCNs)}
\label{sub_sub_sec:fully_connected_network}

are deep neural networks in which an image is provided as input and a mask is generated at the output.
This mask is a binary image (of the same size) where each pixel is classified as iris or not iris.
Basically, we employed the MultiNet~\cite{teichmann_2016_multinet} segmentation decoder without the classification and detection decoders.
The encoder consists of the first $13$ layers of the VGG-$16$ network~\cite{Simonyan_2014_vgg16}. The features extracted from its fifth pooling layer were then used by the segmentation decoder, which follows the \gls*{fcn} architecture~\cite{Shelhamer_2015_CVPR} (see Fig.~\ref{fig:architecture_fcn}).

The fully-connected layers of the VGG-$16$ network were transformed into $1 \times 1$ convolutional layers to produce a low-resolution segmentation.
Then, three transposed convolution layers were used to perform up-sampling.
Finally, high-resolution features were extracted through skip layers from lower layers to improve the up-sampled results.


The segmentation loss function was based on the cross-entropy.
The pre-trained VGG-$16$ weights on ImageNet were used to initialize the encoder, the segmentation decoder, and the transposed convolutional layers.
The training is based on the Adam optimizer algorithm~\cite{kingma_adam_optimizer_2014}, with the following parameters: learning rate of $10^{-5}$, dropout probability of $0.5$, weight decay of $5^{-4}$ and standard deviation of $10^{-4}$ to initialize the skip layers.

\begin{figure}[!htb]
	\begin{center}
		\includegraphics[scale=0.4]{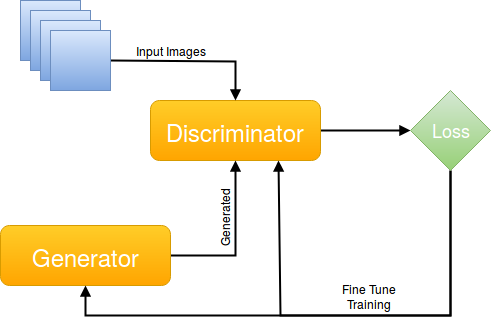}
	\end{center}
	\vspace{-3mm}
	\caption{\gls*{gan} architecture for iris segmentation.}
	\label{fig:architecture_gan}
\end{figure}

\subsubsection{Generative Adversarial Networks (GANs)}
\label{sub_sub_sec:generative_adversarila_network}

are deep neural networks composed by both generator and discriminator networks, pitting one against the other. First, the generator network receives noise as input and generates samples. Then the discriminator network receives samples of training data and those of the generator network, being able to distinguish between the two sources~\cite{goodfellow2014generative}. The \gls*{gan} architecture for iris segmentation is shown in Fig.~\ref{fig:architecture_gan}.

Basically, the generator network learns to produce more realistic samples throughout each iteration, while the discriminator network learns to better distinguish the real and synthetic data.

Isola et al.~\cite{isola2017imagetoimage} presented the \gls*{gan} approach used in this work, which is a \gls*{cgan} able to learn the relation between an image and its label, and from that, generate a variety of image types, which can be employed in various tasks such as photo-generation and semantic segmentation.

%% file: experiments.tex
\section{Experiments}
\label{sec:experiments}

In this section, we present the datasets, evaluation protocol and baselines used in our experiments for comparison of results and discussions.

\subsection{Datasets}
\label{subsec:datasets}

The experiments were carried out on well-known and challenging publicly available iris datasets with both \gls*{nir} and \gls*{vis} images having different sizes and characteristics. An overview of the number of images from each dataset is presented in Table~\ref{tab:overview_datasets}.
The ground truths of the \acs{biosec}, \acs*{casiai3} and \gls*{iitd1} datasets were provided by Hofbauer et al.~\cite{Hofbauer2014}.
In the following, details of the datasets are presented.

\begin{table}[!htb]
	\caption{Overview of the iris datasets used in this work, where (*) means that only part of the dataset was used.}
	\label{tab:overview_datasets}
	\vspace{-1mm}
	\centering
	\resizebox{0.98\linewidth}{!}{
	\begin{tabular}{@{}ccccc@{}}
		\toprule
		\textbf{Dataset}                                             & \textbf{Images} & \textbf{Subjects} & \textbf{Resolution} & \textbf{Wavelength} \\ \midrule
		\acs{biosec}~\cite{Fierrez_2007_biosecDB} (*)                & $400$           & $25$              & $640\times480$      & \gls*{nir}          \\
		\acs{casiai3}~\cite{tan2005casia}                            & $2$,$639$       & $249$             & $320\times280$      & \gls*{nir}          \\
		\acs{casiat4}~\cite{tan2009casia} (*)                        & $1$,$000$       & $50$              & $640\times480$      & \gls*{nir}          \\
		\acs{iitd1}~\cite{Kumar2010}                                 & $2$,$240$       & $224$             & $320\times240$      & \gls*{nir}          \\ 
		\acs{nice1}~\cite{Proenca2012}                               & $945$           & n/a               & $400\times300$      & \gls*{vis}          \\
		\acs{creyeiris}~\cite{Sequeira_2016_Cross-Eyed_database} (*) & $1$,$000$       & $120$             & $400\times300$      & \gls*{vis}          \\
		\acs{miche1}~\cite{demarsico_2015_MICHE_database} (*)        & $1$,$000$       & $75$              & Various             & \gls*{vis}          \\ \bottomrule
	\end{tabular}}
\end{table}

\noindent \textbf{\acs{biosec}}: a multimodal dataset~\cite{Fierrez_2007_biosecDB} containing fingerprint, frontal face and iris images, as well as voice utterances.
The entire dataset has $3$,$200$ \gls*{nir} iris images from $25$ subjects with resolution of $640\times480$ pixels, however, due to the available segmentation masks, we use only  the first $400$ images.

\vspace{1mm}
\noindent \textbf{\gls*{casiai3}}: a dataset~\cite{tan2005casia} with $2$,$639$ \gls*{nir} iris images from $249$ subjects with extremely clear iris texture details and resolution of $320\times280$ pixels, acquired in an indoor environment.

\vspace{1mm}
\noindent \textbf{\gls*{casiat4}}: a dataset~\cite{tan2009casia} containing $20$,$000$ \gls*{nir} images from $1$,$000$ subjects, collected in an indoor environment with different lightings setups.
For our experiments, we manually labeled the first $1$,$000$ images from $50$ subjects.

\vspace{1mm}
\noindent \textbf{\gls*{iitd1}}: a dataset~\cite{Kumar2010} with $2$,$240$ \gls*{nir} images acquired from $224$ subjects between $14$-$55$ years comprising of $176$ males and $48$ females.
All images have a resolution of $320\times240$ pixels and were obtained in an indoor environment.

\vspace{1mm}
\noindent \textbf{\gls*{creyeiris}}:~a~dataset composed of $3$,$840$ images from $120$ subjects~\cite{Sequeira_2016_Cross-Eyed_database}. 
The images were captured with a dual spectrum sensor (\gls*{nir} and \gls*{vis}) and divided into three subsets: iris, masked periocular and ocular images. 
We manually labeled the first $1$,$000$ \gls*{vis} images from the iris subset.

\vspace{1mm}
\noindent \textbf{\gls*{miche1}}: a dataset~\cite{demarsico_2015_MICHE_database} with $3$,$191$ \gls*{vis} images captured from $92$ subjects under uncontrolled settings using three mobile devices: iPhone~5, Galaxy Samsung~IV and Galaxy Tablet~II ($1$,$262$, $1$,$297$ and $632$ images, respectively). The images have resolution of $1536\times2048$, $2320\times4128$ and $640\times480$ pixels, respectively. We used the $569$ ground truth masks made available by Hu et al.~\cite{Hu_2015_MICHE_groundTruth} and labeled another $431$ to complete $1$,$000$ images from $75$ subjects.

\vspace{1mm}
\noindent \textbf{\gls*{nice1}}: 
a subset of the UBIRIS.v2 dataset~\cite{proenca_ubirisv2_2010}.
The \gls*{nice1}~\cite{Proenca2012} subset is composed of $500$ images for training and $500$ for testing. 
However, the test set provided by the organizers of the \gls*{nice1} contest has only $445$ images. The subjects of the test set were not directly specified.

\begin{figure}[!ht]
	\begin{center}
		\subfloat[][]{
			\includegraphics[width=0.45\columnwidth]{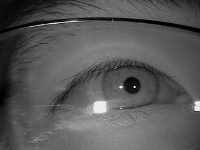}%
			\includegraphics[width=0.45\columnwidth]{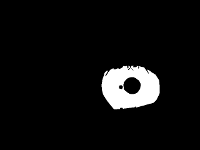}} \; \\[1.5ex]
		\subfloat[][]{
			\includegraphics[width=0.45\columnwidth]{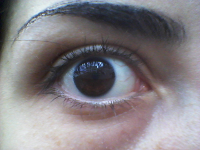}%
			\includegraphics[width=0.45\columnwidth]{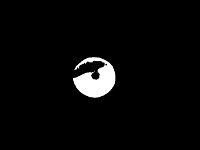}} \;
	\end{center}
	\vspace{-2mm}
	\caption{Two examples of the masks created by us. (a) shows a \gls*{nir} image (\gls*{casiat4}) and (b) a \gls*{vis} image (\gls*{miche1}).}
	\label{fig:masks} 
\end{figure}

Fig.~\ref{fig:masks} shows two samples (\gls*{nir} and \gls*{vis}) of the masks we created.
We sought to eliminate all noise present in the iris, such as reflections and eyelashes.

\subsection{Evaluation protocol}
\label{subsec:metrics}

A pixel-to-pixel comparison between the ground truth (manually labeled) and the algorithm prediction (i.e.,~the mask/segmentation) generate an average segmentation error~$E$ computed as a pixel divergence, given by the exclusive-or logical operator~$\otimes$ (i.e.,~XOR)~\cite{Proenca2012}, denoted by

\begin{equation}
E = \frac{1}{h \times w} \sum_{i} \sum_{j} M_k(i, j) \otimes GT_k(i, j) \, ,
\label{eq:errorSegmentationNICE-1}	
\end{equation}

\noindent where $i$ and $j$ are the coordinates in the mask~$M$ and ground truth~$GT$ images, $h$ and $w$ stand for the height and width of the image, respectively. 
Lower and higher $E$ values represent better and worse results, respectively.  
We also reported the \gls*{f1} measure which is a harmonic average of Precision and Recall~\cite{Jalilian_2017_domain_adaptation}. 

In order to perform a fair evaluation and comparison of the proposed methodologies to the baselines in all datasets, we randomly divided each dataset into two subsets, containing $80\%$ of the images for training and the remainder for evaluation.
The stopping learning criteria was $32$,$000$ iterations.

As suggested in~\cite{teichmann_2016_multinet}, we trained the \gls*{fcn} with $16{,}000$ iterations.
However, we noticed that the more iterations, the better was the model's performance. 
Therefore, we doubled the number of iterations (i.e., $32{,}000$) to ensure a good convergence of the model. 
According to our evaluations, $32{,}000$ iterations were sufficient for all datasets.

\subsection{Benchmarks}
\label{subsec:Baseline}

We selected three baseline frameworks described (and available) in the literature to compare with our approaches with: \gls*{osiris}, \gls*{irisseg} and Haindl \& Krupi\v{c}ka~\cite{Haindl_Krupicka_2015}.

The \textbf{\gls*{osiris}}~\cite{Othman_2016_osiris} framework is composed of four key modules: segmentation, normalization, feature extraction and matching.
Nevertheless, we used only the segmentation module to compare it with our method.
Although the performance of this framework was only reported in datasets with \gls*{nir} images, we applied it on both \gls*{nir} and \gls*{vis} image datasets.
This framework has input parameters such as minimum/maximum iris diameter.
For a fair comparison, we tuned the parameters for each dataset in order to obtain the best results.

The \textbf{\gls*{irisseg}}~\cite{Gangwar_2016_irisSeg} framework was designed specifically for non-ideal irises and is based on adaptive filtering, following a coarse-to-fine strategy.
The authors emphasize that this approach does not require adjustment of parameters for different datasets.
As in \gls*{osiris}, we report the performance of this framework on both \gls*{nir} and \gls*{vis} images.

The \textbf{Haindl \& Krupi\v{c}ka}~\cite{Haindl_Krupicka_2015} framework was used to evaluate the results achieved by the proposed approach on \gls*{vis} datasets.
This method was developed for colored eyes images obtained through mobile devices and used as the baseline in the MICHE-II~\cite{DeMarsico2017} contest.
We did not report the Haindl \& Krupi\v{c}ka~\cite{Haindl_Krupicka_2015} performance on \gls*{nir} images datasets since it was not possible to generate the segmentation masks using the executable provided by the authors.

%% file: results.tex
\section{Results and Discussions}
\label{sec:results}

The experiments were performed using two protocols: the protocol of the \gls*{nice1} contest and the one proposed in Section~\ref{subsec:metrics}.
Moreover, in order to analyze the robustness among sensors from the same environment (i.e., \gls*{nir} or \gls*{vis}) of the proposed \gls*{fcn} and \gls*{gan} approaches, they were training using either all \gls*{nir} or \gls*{vis} image datasets and then evaluated on the same scenario.
Finally, a visual and qualitative analysis showing some good and poor results is performed.

We report the mean \gls*{f1} and $E$ values by averaging the values obtained for each image.
For all the experiments, we also carried out a statistical paired t-test with significance level of $\alpha=0.05$ between pairs of results for the same image, aiming to claim (statistical) significative difference between the results compared.

\subsection{The NICE.I Contest}

The comparison of the results obtained by our approaches and those obtained by the baselines when using the \gls*{nice1} contest protocol is shown in Table~\ref{tab:result_nice_protocol}.
As can be seen, the \gls*{irisseg} and \gls*{osiris} frameworks presented the worst results.
They achieved \gls*{f1} values of $21.76\%$ and $30.70\%$ on the \gls*{nice1} test set, respectively.
These results might be explained because these frameworks were developed for \gls*{nir} images. 
Therefore, their performances are drastically compromised in \gls*{vis} images.
It is noteworthy that the distribution of \gls*{f1} values for both frameworks presented high standard deviation (approximately $\pm 32\%$). 
This occurs because, in some images, the \glspl*{fp} were high in both frameworks, including images that do not have iris, resulting in a very poor segmentation.

\begin{table}[!htb] 
	\centering
	\caption{Iris segmentation results using the \gls*{nice1} contest protocol.}
	\label{tab:result_nice_protocol}
	\resizebox{0.98\linewidth}{!}{
		\begin{tabular}{@{}cccc@{}}
			\toprule
			\, \textbf{Dataset} \, & \textbf{Method} & \textbf{F1 \%} & \textbf{\textit{E} \%} \\ \midrule
			\multirow{4}{*}{\begin{tabular}{@{}c@{}}\, \acs{nice1} \, \\ (\gls*{vis})\end{tabular}}
			& \acs*{osiris}~\cite{Othman_2016_osiris}     &
			$30.70\pm32.00$                   & $08.67\pm06.29$                   \\ 
			& \acs*{irisseg}~\cite{Gangwar_2016_irisSeg}    &
			$21.76\pm32.13$                   & $14.03\pm12.33$                 \\
			& Haindl \& Krupi\v{c}ka~\cite{Haindl_Krupicka_2015}                   &
			$75.54\pm22.93$                   & $03.27\pm04.29$                   \\
			& \textbf{\acrshort*{fcn} Proposed} &
			$\textbf{88.20}\pm\textbf{13.73}$ & $ \textbf{01.05}\pm\textbf{00.86}$ \\  
			& \textbf{\acrshort*{gan} Proposed} &
			$\textbf{91.42}\pm\textbf{03.81}$ & $\textbf{03.09}\pm\textbf{01.76}$
			\\ \bottomrule
		\end{tabular}} \, \,
\end{table}

We expected to obtain good results using the Haindl \& Krupi\v{c}ka~\cite{Haindl_Krupicka_2015} framework, due to the fact that it was developed for \gls*{vis} images and it was used for generating the reference masks (i.e., the ground truth) of the \gls*{miche1} dataset in the recognition contest (MICHE-II).
However, according to our experiments, its performance was not promising, although it obtained better results than \gls*{irisseg} and \gls*{osiris}.

The proposed \gls*{fcn} and \gls*{gan} approaches achieved considerably better  mean values for \gls*{f1} and $E$ metrics than the other approaches. 
We believe that these results were attained due to the discriminating power of the deep learning approaches and also because our models were adjusted (i.e., trained) specifically for each dataset. 
We emphasize that \gls*{osiris} was also adjusted for each dataset.

Although higher standard deviation of \gls*{f1} was presented for the \gls*{fcn}  approach, the paired t-test has shown that the \gls*{gan} approach presented a statistically better \gls*{f1} value, however, the \gls*{fcn} approach has presented a statistically smaller $E$ value.

\subsection{Our protocol}

We trained and tested the \gls*{fcn} and \gls*{gan} approaches on each dataset to compare them with the benchmarks.
Table~\ref{tab:results_proposed_protocol} shows the results obtained when using the proposed evaluation protocol (see Section~\ref{subsec:metrics}).

\begin{table}[!htb]
	\centering
	\caption{Iris segmentation results using the proposed protocol.}
	\label{tab:results_proposed_protocol}
	\resizebox{0.98\linewidth}{!}{
		\begin{tabular}{@{}cccccc@{}}
			\toprule
			\, \textbf{Dataset} \, & \textbf{Method}   
			& \textbf{F1 \%}         & \textbf{\textit{E} \%}          \\ \midrule
			\multirow{4}{*}{\begin{tabular}{@{}c@{}}\acs{biosec} \\ (\gls*{nir})\end{tabular}}
			& \acs*{osiris}~\cite{Othman_2016_osiris}     &
			$92.62\pm03.19$                    & $01.21\pm00.47$                   \\ 
			& \gls*{irisseg}~\cite{Gangwar_2016_irisSeg}           &
			$93.94\pm05.88$                    & $01.06\pm01.20$                   \\
			  & \textbf{\acrshort*{fcn} Proposed} & $\textbf{97.46}\pm\textbf{00.74}$ & $\textbf{00.44}\pm\textbf{00.12}$ \\ 
			  & \textbf{\acrshort*{gan} Proposed} & $\textbf{96.82}\pm\textbf{02.83}$ & $\textbf{00.74}\pm\textbf{01.40}$ \\
			\midrule
			\multirow{4}{*}{\begin{tabular}{@{}c@{}}\acs*{casiai3} \\ (\gls*{nir})\end{tabular}}
			& \acs*{osiris}~\cite{Othman_2016_osiris}     &
			$89.49\pm05.78$                    & $05.35\pm02.40$                   \\
			& \gls*{irisseg}~\cite{Gangwar_2016_irisSeg}           &
			$94.61\pm03.28$                    & $02.85\pm01.62$                   \\
			  & \textbf{\acrshort*{fcn} Proposed} & $\textbf{97.90}\pm\textbf{00.68}$ & $\textbf{01.15}\pm\textbf{00.37}$ \\ 
			& \textbf{\acrshort*{gan} Proposed} &
			$\textbf{96.13}\pm\textbf{05.35}$  & $\textbf{01.45}\pm\textbf{03.71}$ \\
			\midrule
			\multirow{4}{*}{\begin{tabular}{@{}c@{}}\acs*{casiat4} \\ (\gls*{nir})\end{tabular}}
			& \acs*{osiris}~\cite{Othman_2016_osiris}     &
			$87.76\pm08.01$                    & $01.34\pm00.64$                   \\
			& \gls*{irisseg}~\cite{Gangwar_2016_irisSeg}           & 
			$91.39\pm08.13$                    & $00.95\pm00.54$                   \\
			  & \textbf{\acrshort*{fcn} Proposed} & $\textbf{94.42}\pm\textbf{07.54}$ & $\textbf{00.61}\pm\textbf{00.58}$ \\ 
			  & \textbf{\acrshort*{gan} Proposed} & $\textbf{95.38}\pm\textbf{03.72}$ & $\textbf{01.40}\pm\textbf{00.93}$ \\
			\midrule
			\multirow{4}{*}{\begin{tabular}{@{}c@{}}\acs*{iitd1} \\ (\gls*{nir})\end{tabular}}
			& \acs*{osiris}~\cite{Othman_2016_osiris}     & 
			$92.20\pm06.07$                    & $04.37\pm02.69$                   \\
			& \gls*{irisseg}~\cite{Gangwar_2016_irisSeg}           & 
			$94.25\pm03.89$                    & $03.39\pm02.16$                   \\
			  & \textbf{\acrshort*{fcn} Proposed} & $\textbf{97.44}\pm\textbf{01.78}$ & $\textbf{01.48}\pm\textbf{01.01}$ \\ 
			  & \textbf{\acrshort*{gan} Proposed} & $\textbf{95.84}\pm\textbf{04.13}$ & $\textbf{01.33}\pm\textbf{02.65}$ \\
			\midrule
			\midrule
			\multirow{4}{*}{\begin{tabular}{@{}c@{}}\acs*{nice1} \\ (\gls*{vis})\end{tabular}}
			& \acs*{osiris}~\cite{Othman_2016_osiris}     &
			$38.15\pm33.61$                   & $07.92\pm06.20$                   \\
			& \gls*{irisseg}~\cite{Gangwar_2016_irisSeg}           & 
			$28.64\pm35.14$                   & $13.48\pm12.36$                 \\
			& Haindl \& Krupi\v{c}ka~\cite{Haindl_Krupicka_2015}                   & 
			$70.59\pm26.11$                   & $04.72\pm05.87$                   \\
			  & \textbf{\acrshort*{fcn} Proposed} & $\textbf{89.54}\pm\textbf{13.79}$ & $\textbf{01.00}\pm\textbf{00.70}$ \\ 
			& \textbf{\acrshort*{gan} Proposed} & 
			$\textbf{91.12}\pm\textbf{05.08}$ & $\textbf{03.34}\pm\textbf{02.31}$ \\
			\midrule
			\multirow{4}{*}{\begin{tabular}{@{}c@{}}\acs*{creyeiris} \\ (\gls*{vis})\end{tabular}}
			& \acs*{osiris}~\cite{Othman_2016_osiris}     & 
			$46.53\pm29.25$                   & $13.22\pm06.33$                  \\
			& \gls*{irisseg}~\cite{Gangwar_2016_irisSeg}           & 
			$61.72\pm33.55$                   & $10.58\pm10.38$                 \\
			& Haindl \& Krupi\v{c}ka~\cite{Haindl_Krupicka_2015}                   &
			$76.81\pm23.73$                   & $05.69\pm04.58$                   \\
			  & \textbf{\acrshort*{fcn} Proposed} & $\textbf{97.04}\pm\textbf{01.21}$ & $\textbf{00.96}\pm\textbf{00.36}$ \\ 
			& \textbf{\acrshort*{gan} Proposed} & 
			$\textbf{92.61}\pm\textbf{05.86}$  & $\textbf{03.02}\pm\textbf{03.22}$ \\
			\midrule
			\multirow{4}{*}{\begin{tabular}{@{}c@{}}\acs*{miche1} \\ (\gls*{vis})\end{tabular}}
			& \acs*{osiris}~\cite{Othman_2016_osiris}     &
			$33.85\pm35.86$                   & $01.99\pm02.90$                   \\ 
			& \gls*{irisseg}~\cite{Gangwar_2016_irisSeg}           &
			$19.34\pm33.03$                   & $01.90\pm03.37$                   \\
			& Haindl \& Krupi\v{c}ka~\cite{Haindl_Krupicka_2015}                   &
			$63.12\pm33.30$                   & $01.32\pm02.10$                   \\
			  & \textbf{\acrshort*{fcn} Proposed} & $\textbf{83.01}\pm\textbf{19.47}$ & $\textbf{00.37}\pm\textbf{00.43}$ \\ 
			& \textbf{\acrshort*{gan} Proposed} & 
			$\textbf{87.42}\pm\textbf{13.08}$ & $\textbf{03.27}\pm\textbf{03.13}$ \\
			\bottomrule
		\end{tabular}} \,
\end{table}

Remark that both \gls*{irisseg} and \gls*{osiris} frameworks presented good results in \gls*{nir} datasets, always reaching \gls*{f1} values over $90\%$. 
Nonetheless, our proposed approaches presented statistically better \gls*{f1} values for all datasets even in the \gls*{nir} datasets, which are the \gls*{irisseg} and \gls*{osiris} specific image domain.
Observe that there are no results for the approach by Haindl \& Krupi\v{c}ka \cite{Haindl_Krupicka_2015} since it was not developed for \gls*{nir} images.

Looking at \gls*{vis} datasets, the results obtained were slightly worse than in the \gls*{nir} datasets. 
This is because \gls*{vis} images usually have more noise, e.g., reflections. 
The best \gls*{f1} and $E$ values achieved for the \gls*{vis} datasets were achieved by the \gls*{fcn} approach with $97.04\%(\pm01.21)$ and $00.37\%(\pm00.43)$, respectively, in the \gls*{creyeiris} and \gls*{miche1} datasets.

It is worth noting that the \gls*{fcn} approach is the one with the smallest $E$ values in almost all scenarios.
This result can be explained by the fact that the \gls*{fcn} approach took advantage of transfer learning, while the \gls*{gan} approach was trained from scratch.

\begin{table}[!t]
\centering
	\caption{Suitability (bold lines) for \gls*{nir} and \gls*{vis} environments.}
	\label{tab:results_suitability}
		\resizebox{0.98\linewidth}{!}{
		\begin{tabular}{cccccccc}
			\toprule
\textbf{Dataset} & \textbf{Method}  & \textbf{F1 \%}  &\textbf{\textit{E \%}}          \\ \midrule
\multirow{2}{*}{\acs*{biosec}}
                 & \acs*{fcn}   & $97.24\pm00.81$ & $00.58\pm00.30$ \\
                 & \acs*{gan}   & $90.19\pm05.52$ & $02.22\pm01.39$ \\
\multirow{2}{*}{\acs*{casiai3}}
                 & \acs*{fcn}   & $97.43\pm00.74$ & $00.55\pm00.29$ \\
                 & \acs*{gan}   & $97.10\pm01.83$ & $00.75\pm01.10$ \\
\multirow{2}{*}{\acs*{casiat4}}
                 & \acs*{fcn}   & $95.87\pm02.66$ & $01.25\pm00.67$ \\
                 & \acs*{gan}   & $82.65\pm13.98$ & $05.52\pm04.15$ \\
\multirow{2}{*}{\acs*{iitd1}}     
                 & \acs*{fcn}   & $96.47\pm01.56$ & $00.72\pm00.59$ \\ 
                 & \acs*{gan}   & $96.18\pm02.52$ & $01.09\pm01.80$ \\ 
\midrule
\multirow{2}{*}{\acs*{nir}}
                 & \textbf{\acs*{fcn}}   & $\textbf{96.69}\pm\textbf{01.43}$ & $\textbf{00.78}\pm\textbf{00.63}$ \\ 
                 & \textbf{\acs*{gan}}   & $\textbf{94.04}\pm\textbf{07.93}$ & $\textbf{01.72}\pm\textbf{02.69}$ \\ 
\midrule
\midrule
\multirow{2}{*}{\acs*{nice1}}     
                 & \acs*{fcn}   & $90.68\pm14.01$ & $02.67\pm02.04$ \\
                 & \acs*{gan}   & $91.40\pm05.18$ & $01.22\pm00.71$ \\
\multirow{2}{*}{\acs*{creyeiris}}
                 & \acs*{fcn}   & $96.71\pm01.11$ & $01.12\pm00.80$ \\
                 & \acs*{gan}   & $93.21\pm02.30$ & $01.88\pm00.53$ \\
\multirow{2}{*}{\acs*{miche1}}     
                 & \acs*{fcn}   & $88.36\pm11.88$ & $01.90\pm02.20$ \\ 
                 & \acs*{gan}   & $89.49\pm06.76$ & $03.11\pm02.24$ \\ 
\midrule
\multirow{2}{*}{\acs*{vis}}
                 & \textbf{\acs*{fcn}}   & $\textbf{89.56}\pm\textbf{12.36}$ & $\textbf{02.40}\pm\textbf{02.21}$ \\ 
                 & \textbf{\acs*{gan}}   & $\textbf{92.58}\pm\textbf{04.89}$ & $\textbf{02.80}\pm\textbf{02.05}$ \\ \bottomrule
\end{tabular}} \,
\end{table}

\begin{figure*}[!t]
\begin{center}
\subfloat[][\gls*{fcn} $00.31\%$ | $00.85\%$]{
      \includegraphics[width=\figI\linewidth]{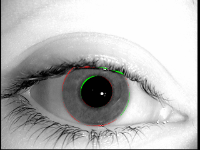}%
      \includegraphics[width=\figI\linewidth]{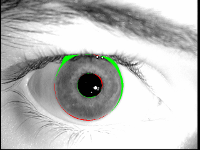}}
\subfloat[][\gls*{gan} $00.27\%$ | $12.61\%$]{
      \includegraphics[width=\figI\linewidth]{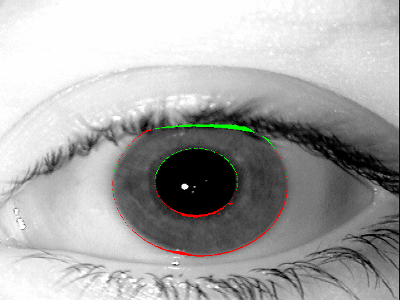}%
      \includegraphics[width=\figI\linewidth]{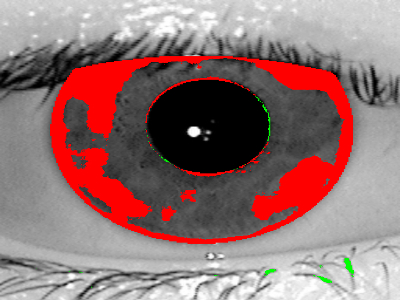}}
\subfloat[][\gls*{fcn} $00.91\%$ | $05.93\%$]{
      \includegraphics[width=\figI\linewidth]{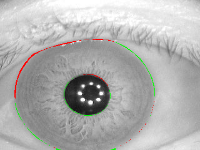}%
      \includegraphics[width=\figI\linewidth]{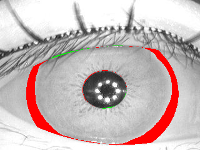}}
\subfloat[][\gls*{gan} $00.43\%$ | $01.51\%$]{
      \includegraphics[width=\figI\linewidth]{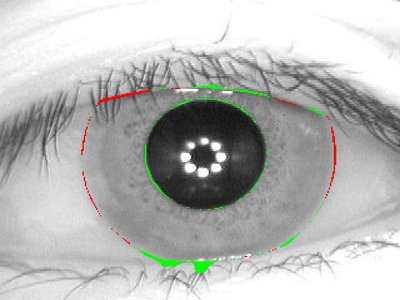}%
      \includegraphics[width=\figI\linewidth]{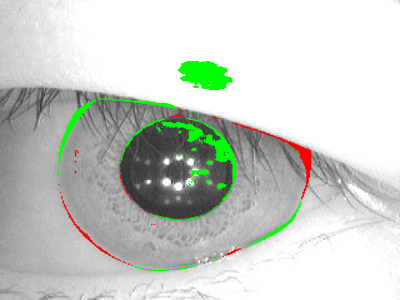}}\\
\subfloat[][\gls*{fcn} $00.52\%$ | $04.57\%$]{
      \includegraphics[width=\figI\linewidth]{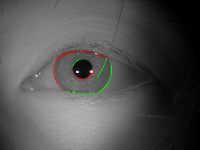}%
      \includegraphics[width=\figI\linewidth]{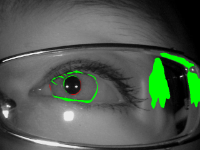}}
\subfloat[][\gls*{gan} $00.84\%$ | $06.30\%$]{
      \includegraphics[width=\figI\linewidth]{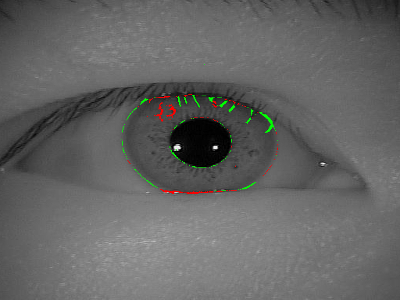}%
      \includegraphics[width=\figI\linewidth]{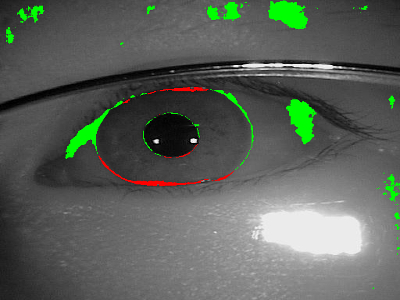}}
\subfloat[][\gls*{fcn} $01.17\%$ | $19.37\%$]{
      \includegraphics[width=\figI\linewidth]{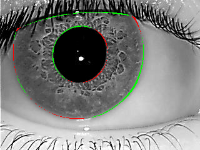}%
      \includegraphics[width=\figI\linewidth]{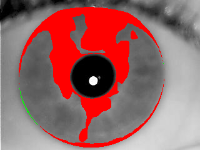}}
\subfloat[][\gls*{gan} $00.56\%$ | $06.60\%$]{
      \includegraphics[width=\figI\linewidth]{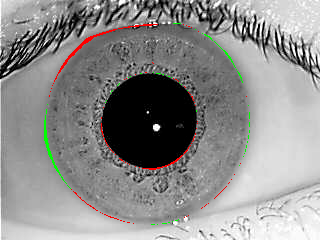}%
      \includegraphics[width=\figI\linewidth]{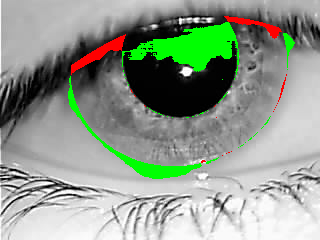}}\\
\subfloat[][\gls*{fcn} $00.95\%$ | $08.28\%$]{
      \includegraphics[width=\figI\linewidth]{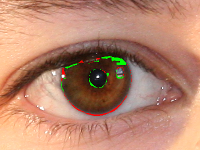}%
      \includegraphics[width=\figI\linewidth]{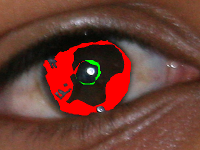}}%
\subfloat[][\gls*{gan} $01.27\%$ | $02.43\%$]{
      \includegraphics[width=\figI\linewidth]{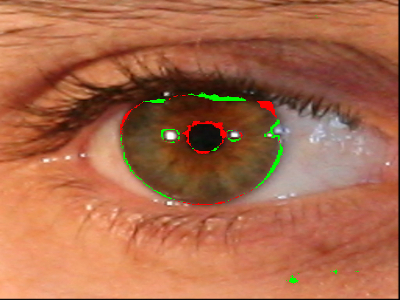}%
      \includegraphics[width=\figI\linewidth]{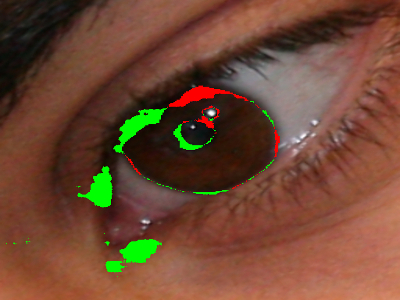}}
\subfloat[][\gls*{fcn} $00.74\%$ | $02.88\%$]{
      \includegraphics[width=\figI\linewidth]{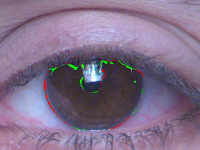}%
      \includegraphics[width=\figI\linewidth]{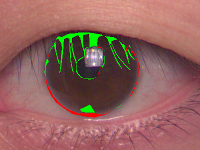}}
\subfloat[][\gls*{gan} $00.72\%$ | $03.61\%$]{
      \includegraphics[width=\figI\linewidth]{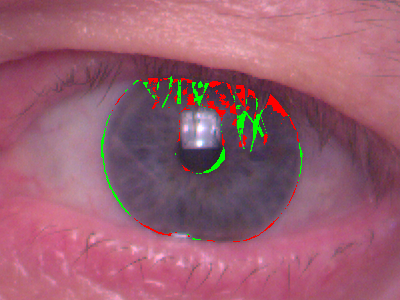}%
      \includegraphics[width=\figI\linewidth]{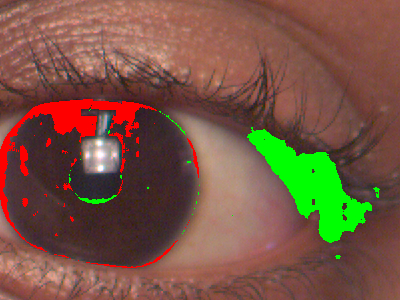}}\\
\subfloat[][\gls*{fcn} $00.42\%$ | $01.82\%$]{
      \includegraphics[width=\figI\linewidth]{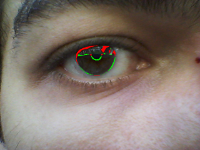}%
      \includegraphics[width=\figI\linewidth]{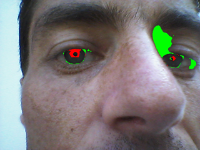}}
\subfloat[][\gls*{gan} $00.57\%$ | $00.96\%$]{
      \includegraphics[width=\figI\linewidth]{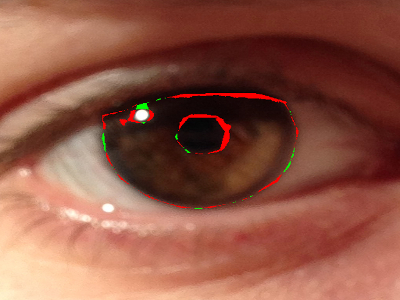}%
      \includegraphics[width=\figI\linewidth]{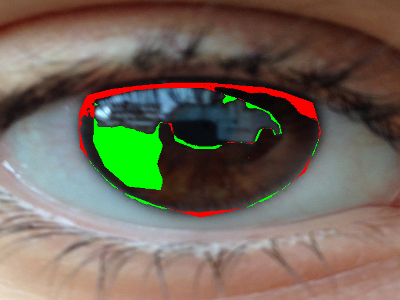}}\\
\end{center}
\caption{\gls*{fcn} and \gls{gan} qualitative results: good (left) and bad (right) results based on the error $E$. Green and red pixels represent the \acrfullpl*{fp} and \acrfullpl*{fn}, respectively. (a)-(b)~\acs{biosec}; (c)-(d)~\acs{casiai3}; (e)-(f)~\acs{casiat4}; (g)-(h)~\acs{iitd1}; (i)-(j)~\acs{nice1}; (k)-(l)~\acs{creyeiris}; (m)-(n)~\acs{miche1}.}
\label{fig:resultadosQualitativos} 
\end{figure*}

\subsection{Suitability and Robustness}

Here, experiments for evaluating the suitability and robustness of the proposed approaches are presented.
By suitability, we expect that models trained with a specific kind of images, i.e. \gls*{nir} or \gls{vis} images, work as well as when training on a specific dataset.
By robustness, we expect that models trained with all kind of images (\gls*{nir} and \gls{vis}) perform as well as when training on a specific dataset.

In summary, the suitability is evaluated by training the models using only \gls*{nir} or \gls*{vis} images (i.e., \gls*{fcn} and \gls*{gan} trained on the \gls*{nir} merged and \gls*{vis} also merged datasets).
The robustness is evaluated by training the models using all images available (\gls*{nir} and \gls*{vis} merged).
The results are presented in Tables~\ref{tab:results_suitability} and~\ref{tab:results_robustness}, respectively.
Note that we report the results of the separate test subsets as well, to facilitate visual comparison between the tables.

\begin{table}[!htb]
	\centering
	\caption{Robustness (bold lines) of the iris segmentation approaches.}
	\label{tab:results_robustness}
	\resizebox{0.98\linewidth}{!}{
		\begin{tabular}{cccccccc}
			\toprule
			\textbf{Dataset} & \textbf{Method} & \textbf{F1 \%}  & \textbf{\textit{E \%}} \\ \midrule
			\multirow{2}{*}{\acs*{biosec}}
			                 & \acs*{fcn}      & $96.57\pm01.14$ & $00.70\pm00.24$        \\
			                 & \acs*{gan}      & $85.48\pm07.63$ & $03.45\pm01.97$        \\
			\multirow{2}{*}{\acs*{casiai3}}
			                 & \acs*{fcn}      & $97.69\pm00.82$ & $00.50\pm00.33$        \\
			                 & \acs*{gan}      & $93.33\pm01.98$ & $00.87\pm00.92$        \\
			\multirow{2}{*}{\acs*{casiat4}}
			                 & \acs*{fcn}      & $95.39\pm03.20$ & $01.46\pm01.12$        \\
			                 & \acs*{gan}      & $85.68\pm12.92$ & $03.98\pm02.80$        \\
			\multirow{2}{*}{\acs*{iitd1}}     
			                 & \acs*{fcn}      & $97.11\pm01.70$ & $00.61\pm00.67$        \\ 
			                 & \acs*{gan}      & $94.99\pm03.88$ & $01.28\pm01.73$        \\ 
			\midrule
			\multirow{2}{*}{\acs*{nir}}
			                 & \acs*{fcn}      & $96.89\pm06,60$ & $00.82\pm00.59$        \\ 
			                 & \acs*{gan}      & $89.87\pm07.93$ & $02.39\pm01.78$        \\ 
			\midrule
			\midrule
			\multirow{2}{*}{\acs*{nice1}}     
			                 & \acs*{fcn}      & $89.25\pm14.06$ & $03.31\pm02.77$        \\
			                 & \acs*{gan}      & $65.56\pm23.32$ & $11.53\pm05.87$        \\
			\multirow{2}{*}{\acs*{creyeiris}}
			                 & \acs*{fcn}      & $96.15\pm01.90$ & $01.38\pm01.16$        \\
			                 & \acs*{gan}      & $88.96\pm08.98$ & $04.57\pm04.63$        \\
			\multirow{2}{*}{\acs*{miche1}}     
			                 & \acs*{fcn}      & $80.49\pm20.65$ & $02.73\pm02.76$        \\ 
			                 & \acs*{gan}      & $61.93\pm24.97$ & $10.95\pm06.22$        \\ 
			\midrule
			\multirow{2}{*}{\acs*{vis}}
			                 & \acs*{fcn}      & $88.63\pm09.15$ & $02.47\pm02.23$        \\ 
			                 & \acs*{gan}      & $72.15\pm19.03$ & $09.01\pm05.54$        \\
			\midrule
			\midrule
			\multirow{2}{*}{All}
			                 & \textbf{\acs*{fcn}}      & $\textbf{94.36}\pm\textbf{09.90}$ & $\textbf{01.26}\pm\textbf{01.73}$        \\ 
			                 & \textbf{\acs*{gan}}      & $\textbf{86.62}\pm\textbf{17.71}$ & $\textbf{04.03}\pm\textbf{05.28}$        \\ \bottomrule
		\end{tabular}} \,
\end{table}

By comparing the values presented in Table~\ref{tab:results_suitability} with those reported in Table~\ref{tab:results_proposed_protocol}, we can observe that the values vary slightly, and thus we can state that the proposed approaches are stable in the suitability scenario.

When comparing the results presented in Table~\ref{tab:results_suitability} and Table~~\ref{tab:results_robustness}, we noticed that the obtained values of $F1$ and $E$ were similar in \gls*{nir} datasets. On the other hand, the performance was considerably lower in \gls*{vis} datasets.
Therefore, the proposed approaches are robust for both \gls*{nir} and \gls*{vis} images.
However, the \gls*{gan} approach presented a decrease in the results, while the \gls*{fcn} obtained little variation.

\subsection{Visual \& Qualitative Analysis}

Here we perform a visual and qualitative analysis.
First, in Fig.~\ref{fig:resultadosQualitativos}, we show poor and well-performed iris segmentation results obtained in each dataset by the \gls*{fcn} and \gls*{gan} approaches.
Some images were poorly segmented, thus explaining the high standard deviations obtained.

\begin{figure}[!htb]
\captionsetup[subfigure]{labelformat=empty}
\begin{center}
\subfloat[][\acs*{osiris}]{
      \includegraphics[width=0.235\linewidth]{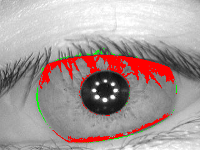}}
\subfloat[][IrisSeg]{
      \includegraphics[width=0.235\linewidth]{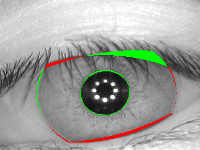}}
\subfloat[][\gls*{fcn}]{
      \includegraphics[width=0.235\linewidth]{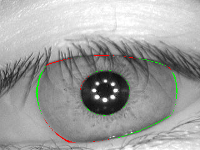}}
\subfloat[][\gls*{gan}]{
      \includegraphics[width=0.235\linewidth]{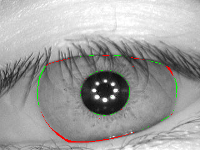}}
\\
\subfloat[][\acs*{osiris}]{
      \includegraphics[width=0.185\linewidth]{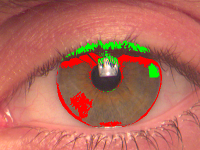}}
\subfloat[][IrisSeg]{
      \includegraphics[width=0.185\linewidth]{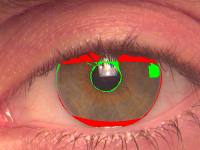}}
\subfloat[][Haindl~\& Krupi\v{c}ka~\cite{Haindl_Krupicka_2015}]{
      \includegraphics[width=0.185\linewidth]{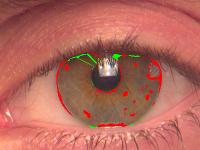}} 
\subfloat[][ \gls*{fcn}]{
      \includegraphics[width=0.185\linewidth]{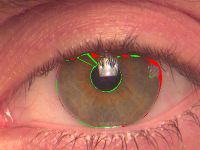}}
\subfloat[][\gls*{gan}]{
      \includegraphics[width=0.185\linewidth]{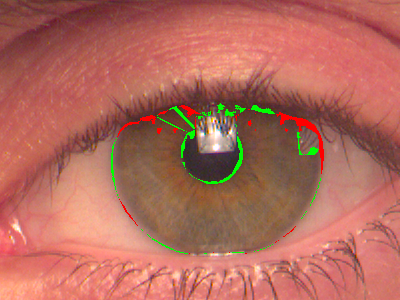}}\\
\end{center}
\caption{Qualitative results achieved by the \gls*{fcn}, \gls*{gan} and baselines. Green and red pixels represent the \gls*{fp} and \gls*{fn}, respectively. The first and second rows correspond, respectively, to images from the \gls*{casiai3} and \gls*{creyeiris} datasets.}
\label{fig:comparacao_qualitativa} 
\end{figure}

Then, in Fig.~\ref{fig:comparacao_qualitativa}, we show iris segmentation performed by both the \gls*{fcn} and \gls*{gan} approaches, as well as the baselines.
We only show one image from each the \gls*{casiai3} and \gls*{creyeiris} datasets due to lack of space.

We particularly chose images where all methods perform fairly well and also where our methods performed better, which is the case in most situations. 
One can observe that our approach performed better in both \gls*{nir} and \gls*{vis} images.

%% file: conclusion.tex
\section{Conclusion}
\label{sec:conclusion}

This work presented two approaches (\gls*{fcn} and \gls{gan}) for robust iris segmentation in \gls*{nir} and \gls*{vis} images in both cooperative and non-cooperative environments.
The proposed approaches were compared with three baselines methods and reported better results in all test cases.
The transfer learning for each domain (or dataset) was essential to achieve outstanding results since the number of images for training the \gls*{fcn} is relatively small.
Therefore, the use of pre-trained models from other datasets brings excellent benefits in learning deep networks.
Moreover, specific data augmentation techniques can be applied for improving the performance of the \gls*{gan} approach.

We also labeled more than $2$,$000$ images for iris segmentation. 
These masks (manually labeled) are publicly available to the research community, assisting the development and evaluation of new iris segmentation approaches.

Despite the outstanding results, our approach presented high standard deviation rates in some datasets. 
Therefore, as future work we intend to (i)~evaluate the impact of performing the segmentation in two steps, that is, first perform iris detection and then segment the iris in the detected patch; (ii)~create a post-processing stage to refine the prediction, since many images have minor errors (especially at the limbus); (iii)~first classify the sensor or image type and then segment each image with a specific and tailored convolutional network model, in order to design a general approach.